\documentclass[10pt,twocolumn,letterpaper]{article}

\usepackage[pagenumbers]{cvpr} 

\usepackage{graphicx}
\usepackage{times}
\usepackage{helvet}
\usepackage{courier}
\usepackage{amsmath}
\usepackage{algorithm}
\usepackage{csquotes} 
\usepackage{paralist}
\usepackage{amssymb}
\usepackage{indentfirst}
\usepackage{float}
\usepackage{multirow}
\usepackage{cite}
\usepackage{textcomp,booktabs}
\usepackage{amssymb}
\usepackage{pifont}
\usepackage[misc]{ifsym}

\newcommand{\cmark}{{\ding{51}}}     
\newcommand{\xmark}{{\ding{55}}} 

\usepackage{makecell}
\usepackage{xcolor}
\definecolor{citecolor}{HTML}{0071bc} 

\usepackage{mathrsfs}
\usepackage[colorlinks, linkcolor=red,  anchorcolor=blue, citecolor=citecolor]{hyperref}
\usepackage{CJKutf8} 
\usepackage{algpseudocode}

\usepackage{soul}
\usepackage{tcolorbox}

\usepackage[capitalize]{cleveref}
\crefname{section}{Sec.}{Secs.}
\Crefname{section}{Section}{Sections}
\Crefname{table}{Table}{Tables}
\crefname{table}{Tab.}{Tabs.}

\def\paperID{10107}      
\def\confName{ICCV}
\def\confYear{2025}

\title{ Sign Language Translation using Frame and Event Stream: Benchmark Dataset and Algorithms }  

\author{
  Xiao Wang$^{1}$, Yuehang Li$^{1}$, Fuling Wang$^{1}$, Bo Jiang$^{1}$ \thanks{Corresponding Author: Bo Jiang}, Yaowei Wang$^{2,3}$ \\
  Yonghong Tian$^{3,4,5}$, Jin Tang$^{1}$, Bin Luo$^{1}$ \\
  ${^1}${School of Computer Science and Technology, Anhui University, Hefei, China} \\
  ${^2}${Harbin Institute of Technology, Shenzhen, China} \\
  ${^3}${Peng Cheng Laboratory, Shenzhen, China} \\
  ${^4}${National Key Laboratory for Multimedia Information Processing, Peking University, China} \\
  ${^5}${School of Electronic and Computer Engineering, Shenzhen Graduate School, Peking University, China} \\
  \textit{\{xiaowang, jiangbo, tangjin, luobin\}@ahu.edu.cn}, \textit{e23201112@stu.ahu.edu.cn}, \\
  \textit{18870722722@163.com}, \textit{wangyw@pcl.ac.cn}, \textit{yhtian@pku.edu.cn}
}

\begin{document}
\maketitle

\begin{abstract}
Accurate sign language understanding serves as a crucial communication channel for individuals with disabilities. Current sign language translation algorithms predominantly rely on RGB frames, which may be limited by fixed frame rates, variable lighting conditions, and motion blur caused by rapid hand movements. Inspired by the recent successful application of event cameras in other fields, we propose to leverage event streams to assist RGB cameras in capturing gesture data, addressing the various challenges mentioned above. Specifically, we first collect a large-scale RGB-Event sign language translation dataset using the DVS346 camera, termed VECSL, which contains 15,676 RGB-Event samples, 15,191 glosses, and covers 2,568 Chinese characters. These samples were gathered across a diverse range of indoor and outdoor environments, capturing multiple viewing angles, varying light intensities, and different camera motions. Due to the absence of benchmark algorithms for comparison in this new task, we retrained and evaluated multiple state-of-the-art SLT algorithms, and believe that this benchmark can effectively support subsequent related research. Additionally, we propose a novel RGB-Event sign language translation framework (i.e., M$^2$-SLT) that incorporates fine-grained micro-sign and coarse-grained macro-sign retrieval, achieving state-of-the-art results on the proposed dataset. 
Both the source code and dataset will be released on \url{https://github.com/Event-AHU/OpenESL}. 
\end{abstract}


\section{Introduction}

Sign language is the principal means of communication utilized by deaf individuals in daily life. It serves as a vital bridge that enables them to express their thoughts, emotions, and ideas effectively within their own community and with the wider society. The communication barriers between deaf and hearing people remain a significant challenge. In order to bridge this gap and enable seamless interaction, Sign Language Translation (SLT) has emerged as a highly promising direction of research.

Typically, most existing sign language translation algorithms are based on videos captured by traditional frame cameras (25-30 FPS) as input. 
The model first needs to encode them into visual features, and then utilize a text decoder to obtain the sign language translation. 
With the development of deep learning technologies, especially large language models like ChatGPT~\footnote{\url{https://openai.com/index/chatgpt/}}, machines' visual perception and text generation capabilities have significantly improved. 
However, existing SLT datasets~\cite{zhou2021improving, camgoz2018neural} are typically collected in constrained environments using traditional frame-based cameras, such as laboratory settings or CCTV systems. It is well known that RGB cameras are highly susceptible to the effects of low illumination, overexposure, complex backgrounds, and motion blur. 
Consequently, their fixed backgrounds, limited angles, and controlled lighting conditions differ significantly from the complex and dynamic scenarios encountered in real-life settings. 
It is difficult to ensure that SLT algorithms can maintain high performance in real-life scenarios. The scarcity of real-world outdoor scene data has now become a key bottleneck hindering advancements in sign language translation.

\begin{figure}
\centering
\includegraphics[width=1\linewidth]{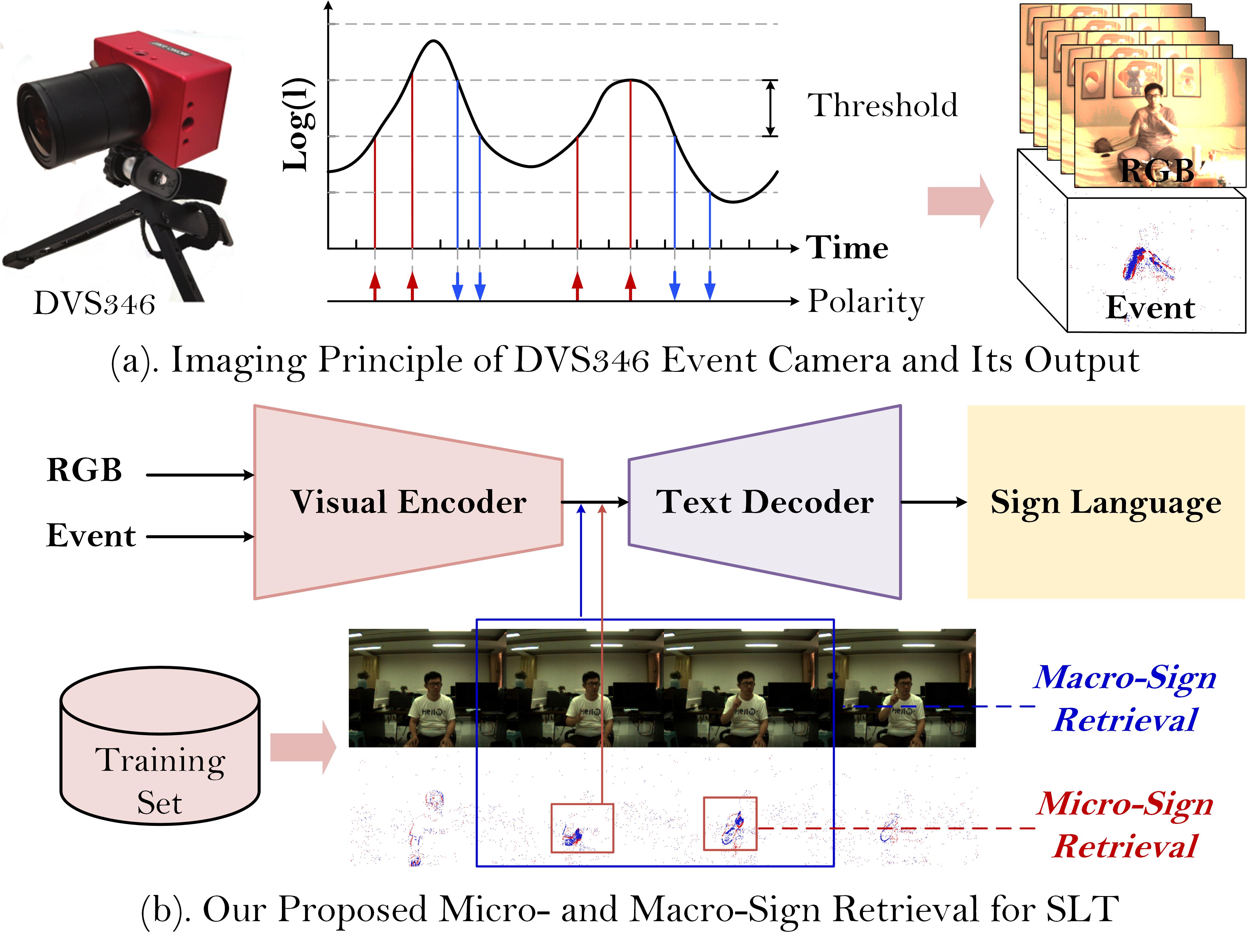}
\caption{
The imaging principle of event cameras and an overview schematic of our proposed M$^2$-SLT.} 
\label{fig:firstIMG}
\end{figure}

To address the issues caused by RGB cameras, some researchers resort to the event cameras for sign language translation~\cite{Wang2021Event-Based, zhang2024evsign, jiang2024evcslr}. 
Compared to traditional frame-based cameras, event cameras offer a \textit{wider dynamic range}, \textit{lower energy consumption}, and a \textit{denser temporal resolution}, enabling more sensitive capture of moving objects. 
As shown in Fig.~\ref{fig:firstIMG} (a), when the variation of light intensity exceeds a specific threshold, the event camera outputs a corresponding event point ($x, y, t, p$), where ($x, y$) represents the pixel's spatial position, and $t$ and $p$ denote the timestamp and polarity, respectively (e.g., red and blue event points triggered by increases and decreases in brightness represent ON and OFF events). 
Using event cameras alone can offer superior privacy protection for humans; however, SLT algorithms based on event cameras still cannot achieve performance comparable to RGB cameras.



In this paper, we formally propose a multi-modal framework that integrates RGB frames and event streams to achieve high-precision, robust sign language translation. We first propose a large-scale multi-modal sign language translation dataset, termed \textbf{VECSL}, to better advance research in this direction. It is collected using the DVS346 event camera which outputs spatial-temporal aligned RGB frames and event streams. VECSL dataset contains 15,676 RGB-Event samples (the resolution is $346 \times 260$), 15,191 glosses, and covers 2,568 Chinese characters. It fully reflects the key challenges in the practical scenarios, including diverse indoor and outdoor environments, multiple angles, different illuminations and camera motions. We split them into the training, validation, and testing subsets, which contain 14108, 744, and 824 samples, respectively. In addition, we also build a benchmark by re-training and reporting the experimental results of multiple strong sign language translation algorithms. We believe the construction of this benchmark dataset can provide a strong foundation for multi-modal sign language translation tasks.

Based on our newly proposed benchmark dataset, we introduce a novel micro-sign and macro-sign retrieval-guided framework for sign language translation, termed \textbf{M$^2$-SLT}, as illustrated in Fig.~\ref{fig:firstIMG} (b). The key insight of our framework is inspired by the dual-process mechanism of human sign language comprehension. Specifically, when observing a signer performing sign language, humans tend to focus on the patterns of hand movement changes (analogous to micro-sign retrieval in our framework) while simultaneously leveraging their existing linguistic knowledge to compare and process the overall sequence of the signing process (analogous to macro-sign retrieval). In more detail, given RGB frames and event streams, we first utilize a parameter-shared backbone network to extract their respective visual features. Subsequently, the Micro-Sign Retrieval (MiR) module is employed to enhance and fuse multi-modal features through a shared memory pool. Additionally, the Macro-Sign Retrieval (MaR) module is introduced to retrieve representative features from the training set. The resulting RGB and event stream feature representations are further enhanced using a Hopfield network and combined with those refined by the MiR module. Finally, the text decoder mBART~\cite{Liu2020Multilingual} is utilized for sign language translation, integrating the enriched multi-modal representations to achieve accurate and robust translation. Extensive experiments on VECSL dataset fully validated the effectiveness of our proposed framework for SLT. 

To sum up, we draw the main contributions of this work as the following three aspects: 

1). We propose \textit{a large-scale dataset} for RGB-Event based sign language translation, termed VECSL. 
It fully reflects the challenges/attributes like diverse indoor and outdoor environments, multiple angles, different illuminations, and camera motions. 

2). We propose \textit{a new sign language translation framework} based on micro-sign and macro-sign retrieval, termed M$^2$-SLT. 
These two modules effectively reflect the human focus on the motion of sign language gestures and the associative memory of certain actions in the mind when comprehending sign language.

3). We build \textit{a benchmark} for the VECSL dataset by re-training and reporting the experimental results of multiple strong SLT algorithms. 
We believe the construction of this benchmark can well inspire subsequent research and provide a solid foundation for future works to compare.

\section{Related Work} 

In this section, we review the related works on Sign Language Translation, Event-based Vision, and Retrieval-Augmented Generation. 

\noindent $\bullet$ \textbf{Sign Language Translation.~} 
Sign Language Translation (SLT) is recognized as one of the most socially impactful applications of computer vision. The integration of these fields not only drives technological advancements in gesture recognition and multi-modal learning~\cite{camgoz2018neural, camgoz2020sign, guo2018hierarchical} but also facilitates equitable communication access for individuals with disabilities.
Some researchers~\cite{zhou2021improving, chen2022simple, zhang2025scaling} mitigate data scarcity by enhancing diversity through back-translation augmentation, cross-modal transfer learning, and large-scale pre-training strategies. Building upon this, some scholars~\cite{yin2022mlslt, Chaudhary2023SignNet} propose a dynamic routing mechanism integrated with a bidirectional translation architecture, enabling multilingual mutual translation and bidirectional text-sign language conversion to improve system generalizability. Zhou et al.~\cite{zhou2023GFSLT} and Ye et al.~\cite{ye2025improving} optimize cross-modal feature alignment in latent space through vision-language pre-training and contrastive learning paradigms, effectively eliminating reliance on annotated data while enhancing translation robustness. 
To advance spatiotemporal modeling capabilities, some researchers~\cite{yao2023IP-SLT, camgoz2020multi} design an iterative prototype optimization framework coupled with a multi-channel Transformer architecture, specifically tailored for continuous sign language movement analysis. 
Gong et al.~\cite{gong2024llms} pioneers a novel paradigm that encodes sign language videos into discrete linguistic representations, achieving end-to-end annotation-free translation via integration with large language models. 
Different from these works, we propose the RGB-Event sign language translation task to achieve high-performance SLT.

\noindent $\bullet$ \textbf{Event-based Vision.~} 
Event cameras offer high temporal resolution, low latency, and a wide dynamic range, effectively reducing motion blur and adapting to extreme lighting conditions. Several studies~\cite{gehrig2023rvt, hamann2024reTAG, luo2024EEMFlow, wan2023rpeflow, peng2023get} have leveraged these advantages, making significant progress in various event-based vision applications.
Yang et al.~\cite{yang2023event} proposed a self-supervised pre-training method for event cameras that enhances downstream task performance. 
AE-NeRF~\cite{feng2025aeNeRF} advances event-based NeRF reconstruction under challenging conditions, 
while Wang et al.~\cite{wang2024MvHeatDet} contribute the EvDET200K dataset for multi-category event-based detection. 
LEOD~\cite{wu2024leod} demonstrates self-training strategies for event camera object detection, and Zubic et al.~\cite{zubic2024state} explore state-space models to improve event camera generalization.
Notably, EvSign~\cite{zhang2024evsign} and EvCSLR~\cite{jiang2024evcslr} pioneer event-based sign language translation by leveraging motion blur reduction and real-time processing. Their findings reveal event cameras' unique potential in capturing high-speed gestures under complex lighting conditions while maintaining computational efficiency and privacy.
Inspired by these advances, our work leverages event streams to complement RGB data in sign language translation, thereby effectively addressing the limitations of conventional frame-based approaches.


\noindent $\bullet$ \textbf{Retrieval-Augmented Generation.~} 
Retrieval-Augmented Generation (RAG) integrates information retrieval with text generation by leveraging external knowledge bases to enhance the accuracy, real-time performance, and interpretability of generative models~\cite{jiang2023activeRAG,wang2024searchingRAG,cheng2025xrag}.
Some researchers~\cite{lewis2020retrieval,ram2023context} propose hybrid architectures integrating parametric language models with non-parametric knowledge bases via  RAG, significantly enhancing factual accuracy and semantic diversity in open-domain text generation. 
Other scholars~\cite{jiang2023activeRAG,zhang2023remodiffuse} design proactive retrieval mechanisms that dynamically trigger retrieval by anticipating future content states, effectively improving semantic coherence in long-text generation and complex motion synthesis. 
A majority of researchers~\cite{blattmann2022retrieval,long2022retrieval,liu2023learning} introduce cross-modal retrieval-augmented frameworks that integrate lightweight parametric models with external databases, substantially improving recognition performance on long-tailed visual data and cross-domain generalization. 
Cheng et al.~\cite{cheng2023lift} proposes a self-memory iterative optimization framework that constructs and refines dynamically updated memory pools, reducing reliance on external knowledge sources and enabling progressive enhancement of generation quality. 
Wang et al.~\cite{wang2024r2gencsr,wang2025ammrg} develops a domain-specific RAG method for medical report generation by integrating clinical knowledge bases with parametric models, markedly improving clinical accuracy and semantic coherence.
By integrating RAG's external knowledge retrieval with high-temporal-resolution data from event cameras, sign language translation can effectively overcome information scarcity and motion blur, enabling real-time, accurate translation.



\section{Methodology}  

\begin{figure*}[h]
\centering
\includegraphics[width=\linewidth]{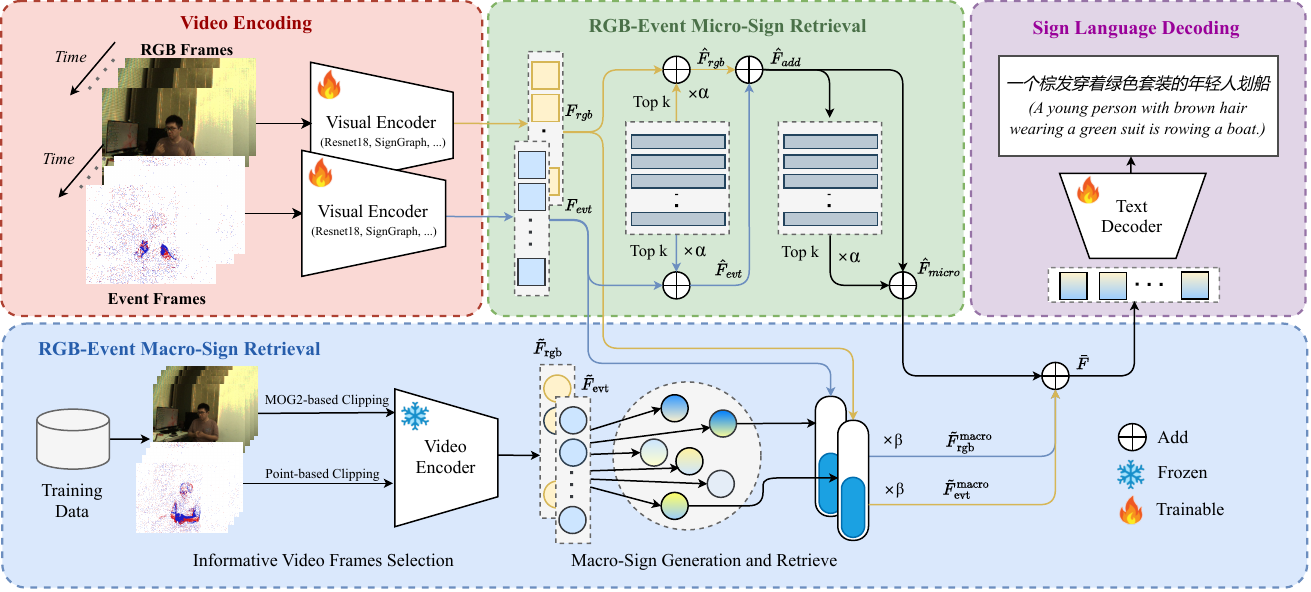}
\caption{An overview of our proposed framework for sign language translation, i.e., M$^2$-SLT. It consists of four main components, including RGB and event encoding, micro-sign retrieval (MiR), macro-sign retrieval (MaR), and sign language decoding.}
\label{fig:framework}
\end{figure*}

\subsection{Overview}  
As shown in Fig.~\ref{fig:framework}, our framework takes RGB frames and event streams as input and employs parameter-shared backbone networks (such as ResNet~\cite{he2016resnet}, SignGraph~\cite{gan2024signgraph}) to extract their respective visual features. Then, the MiR (Micro-Sign Retrieval) is adopted to enhance and fuse the multi-modal features via shared memory pools. On the other hand, we propose the MaR (Macro-Sign Retrieval) module to retrieve contextual samples for enriching the multi-modal representation. Specifically, we first de-duplicate the samples in the training set and convert them into feature embeddings. These features are then clustered using DBSCAN~\cite{ester1996dbscan}, yielding 123 representative prototype features. We employ the Hopfield network to perform retrieval-based enhancement among these features. The resulting RGB and event stream feature representations are subsequently combined with the features enhanced by the MiR module. Finally, we adopt a text decoder mBART~\cite{Liu2020Multilingual} for sign language translation.

\subsection{Network Architecture} 
In this section, we will introduce our framework from the four key modules, i.e., the Input Encoding Network, RGB-Event Micro-Sign Retrieval, RGB-Event Macro-Sign Retrieval, and Sign Language Decoding Network.

\subsubsection{Input Encoding Network}  
In this work, the event streams can be represented as $\mathcal{E} = \{e_1, e_2, ..., e_M\}$, where $e_i = [x, y, t, p]$ denotes the $i^{th}$ event point in the stream. Here, the ($x, y$) is the spatial coordinate, $t$ and $p$ denote the time stamp and polarity (+1 or -1) respectively. To better train existing benchmark algorithms that use frame-based input, we convert event streams into frames $X_{evt} \in \mathbb{R}^{T \times H \times W \times 3}$ for subsequent processing by spatio-temporally aligning with the RGB frames $X_{rgb} \in \mathbb{R}^{T \times H \times W \times 3}$, where $T$ is the number of frames for each RGB/Event sample, and $H, W$ are hight and width of input frames.

Given the aligned RGB-Event frames, we adopt a visual encoder network (the ResNet18~\cite{he2016resnet} and SignGraph~\cite{gan2024signgraph} are used in our experiments) to extract their features and get $F_{rgb} \in \mathbb{R}^{L \times D}$ and $F_{evt} \in \mathbb{R}^{L \times D}$, where the $L$ is sequence length, $D$ is feature dimension. For the SignGraph, it is a graph convolutional network that is adapted for spatial and temporal visual feature learning. The RGB features capture the holistic appearance and gradual motion, while the event features emphasize high-frequency movements and transient gestures. Then, we feed them into the MiR and MaR modules simultaneously for multi-modal feature fusion and enhancement.

\subsubsection{RGB-Event Micro-Sign Retrieval}  

After we get the RGB and Event features $F_{rgb} \in \mathbb{R}^{L \times D}$ and $F_{evt} \in \mathbb{R}^{L \times D}$, we adopt a Micro-Sign Retrieval (MiR) module to further enhance single-modality and multi-modal feature representations. Inspired by the observation that micro-sign (e.g., finger articulations, transient motions) constitute fundamental semantic units in sign language, we design a memory-augmented architecture to retrieve and amplify these critical patterns. As shown in Fig.~\ref{fig:framework}, the MiR module contains randomly initialized and learnable \textit{micro-sign memory} $M \in \mathbb{R}^{P \times \hat{L}} = \{ m_1, m_2, ..., m_{P} \}$ representing common patterns. Here, $P$ and $\hat{L}$ are the number of memory vectors and feature dimension, respectively.

\begin{figure*}
\centering
\small
\includegraphics[width=\linewidth]{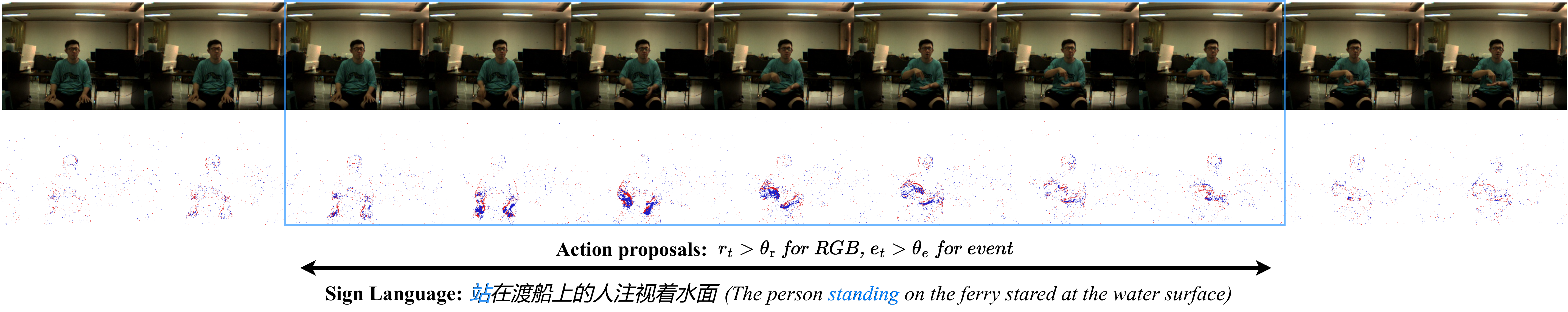}
\caption{An illustration of informative video frames selection in RGB-Event Macro-Sign Retrieval (MaR).}
\label{fig:visualize_clip}
\end{figure*}

Let's first take the RGB modality as an example to illustrate the basic process of the Micro-Sign Retrieval enhancement strategy. For the RGB feature $F_{rgb}$, we project it into a latent space $Z \in \mathbb{R}^{L \times d}$ via a MLP (Multi-Layer Perceptron) $\phi_{\text{Enc}}(\cdot)$, where $d$ compresses the dimensionality to focus on salient attributes. Then, we compute its cosine similarity with the micro-sign memory $M$ and obtain the similarity matrix $S \in \mathbb{R}^{L \times \hat{L}}$, 
\begin{equation}
\small 
S_{i,j} = \frac{Z_i \cdot M_j}{\|Z_i\| \|M_j\|}, \quad \forall i \in [1,  L], \forall j \in [1, \hat{L}]
\label{eq:micro_equation}
\end{equation}
A sparse top-$k$ selection strategy is employed to enable dynamic pattern retrieval. We average these features and project them to align with the same dimensionality as the input features using an additional MLP $\phi_{\text{Dec}}$. Subsequently, we add the raw RGB feature, i.e.,  
\begin{equation}
\hat{F}_{rgb} = F_{rgb} + \alpha \cdot \phi_{\text{Dec}}(\frac{1}{k}\sum_{i=1}^k M^{(i)}) \label{eq:micro_equation}
\end{equation}
where $\alpha$ is a learnable scaling parameter. Similar operations can be conducted for the event stream branch and get the enhanced event features $\hat{F}_{evt} \in \mathbb{R}^{L \times D}$.

After that, the modality-specific enhanced features $\hat{F}_{rgb}$ and $\hat{F}_{evt}$ from Eq.~\eqref{eq:micro_equation} are first aggregated through element-wise addition, i.e., $\hat{F}_{add} = \hat{F}_{rgb} + \hat{F}_{evt}$. 
Then, we project them into the memory space for progressive refinement of micro-sign representations: 
\begin{equation}
\hat{F}_{micro} = \hat{F}_{add} + \alpha \cdot \phi_{\text{Dec}}(\frac{1}{k}\sum_{i=1}^{k} M^{(i)})
\end{equation}
Our proposed memory-guided retrieval mechanism suppresses noise by reinforcing micro-sign patterns learned from training data. 
The shared micro-sign memory enables cross-modal alignment of similar gestures appearing in different modalities and the lightweight architecture ensures efficient computation without heavy parameter overhead.




\subsubsection{RGB-Event Macro-Sign Retrieval}  

While micro-sign retrieval focuses on fine-grained gesture patterns, we also propose a Macro-Sign Retrieval (MaR) module to model coarse-grained semantic units (e.g., complete lexical signs or multi-word phrases). 
MaR contains two main stages: 1) Informative Video Frames Selection; 2) Macro-Sign Generation and Retrieval.

\noindent $\bullet$ \textbf{Informative Video Frames Selection.~} 
Since the start and end times of video recording typically precede and extend beyond the signer's effective motion period, the raw data contains some redundancy. To address this, we perform cropping operations on both the RGB frames and event streams to extract truly informative data, thereby reducing computational overhead for subsequent model processing.
For the RGB frames, we compute the motion intensity $ r_t $ at each timestep through foreground pixel counting using background subtraction. Action intervals are identified when the motion intensity exceeds a predefined threshold $ \theta_{\text{r}} $, with the additional constraint that this elevated state must persist for at least $ \alpha_{\min} $ consecutive frames to ensure temporal consistency. For event streams, we calculate the event count $ e_t $ per frame and employ an adaptive thresholding mechanism $ \theta_e = \frac{1}{T}\sum_{t=1}^{T+1}e_t $, dynamically adjusted based on temporal statistics. Action proposals $ \mathcal{A}_{\text{rgb}} $ and $ \mathcal{A}_{\text{evt}} $ are subsequently generated when the respective modality's measurements ($ r_t $ for RGB or $ e_t $ for event streams) surpass their corresponding thresholds continuously over $ \alpha_{\min} $ frames, ensuring synchronized multi-modal temporal localization.

To ensure temporal consistency, we unify the action proposals by prioritizing event-based segmentation due to its precise temporal resolution. The merged segments \(\mathcal{P} = \{p_i\}_{i=1}^m\) (where \(p_i = [s_i, e_i]\) denotes start/end indices) are constructed by first aligning RGB proposals \(\mathcal{A}_{\text{rgb}}\) to event-based anchors \(\mathcal{A}_{\text{evt}}\). This alignment ensures that both modalities share a coherent set of temporal boundaries, with \(\mathcal{P}\) serving as the synchronized reference for downstream tasks. Event-driven anchors dominate boundary adjustments due to their microsecond-level sensitivity, while RGB proposals refine the merged segments through validation.
An illustration of this video frames selection can be found in Fig.~\ref{fig:visualize_clip}.

\noindent $\bullet$ \textbf{Macro-Sign Generation and Retrieval.~} 
For each aligned segment $p_i \in \mathcal{P}$, we extract spatio-temporal features $\tilde{F}_{\text{rgb}}$ and $\tilde{F}_{\text{evt}}$ using sliding windows (stride $\tau=2$, window size $k=8$) with a pretrained video encoder. To address memory constraints when processing large-scale training sets, we employ a two-stage compression strategy. First, cross-modal features are aggregated through element-wise averaging, i.e., $\tilde{F}_{\text{macro}} = \frac{1}{2}(\tilde{F}_{\text{evt}} + \tilde{F}_{\text{rgb}})$. Then, we apply DBSCAN clustering~\cite{ester1996density} with adaptive $\epsilon$ thresholds to extract representative prototypes: 
\begin{align}
\mathcal{C} &= \mathop{\text{DBSCAN}}\limits_{\epsilon=f(\mathcal{P})}(\tilde{F}_{\text{macro}}) \label{eq:cluster} \\
\tilde{F}_{\text{*}}^{\text{macro}} &= \{\mu_c | \mu_c = \mathbb{E}[F_i], \forall F_i \in c\}_{c=1}^C \label{eq:prototype}
\end{align}
where $\mathcal{C}$ contains $C$ clusters and $\mu_c$ represents cluster centroids. Finally, compressed features undergo Hopfield network~\cite{ramsauer2020hopfield}-based retrieval:
\begin{align}
\small 
\tilde{F}_{\text{rgb}}^{\text{macro}} &= {F}_{\text{rgb}}+ \beta \cdot \phi_{\text{Dec}}(\phi_{\text{Hopfield}}(\phi_{\text{Enc}}({F}_{\text{rgb}}); \tilde{F}_{\text{*}}^{\text{macro}})) \label{eq:hop_rgb} \\
\tilde{F}_{\text{evt}}^{\text{macro}} &= {F}_{\text{evt}}+ \beta \cdot \phi_{\text{Dec}}(\phi_{\text{Hopfield}}(\phi_{\text{Enc}}({F}_{\text{evt}}); \tilde{F}_{\text{*}}^{\text{macro}})) \label{eq:hop_evt} 
\end{align}
where $\beta$ is a learnable scaling parameter and the Hopfield network $\phi_{\text{Hopfield}}$ utilize the prototypes $\tilde{F}_{\text{*}}^{\text{macro}}$ as memory patterns.

\subsubsection{Sign Language Decoding}  
The decoding stage synthesizes enhanced spatiotemporal representations from both micro and macro perspectives through hierarchical feature fusion. As shown in Fig. \ref{fig:framework}, we aggregate the refined features via element-wise summation to preserve complementary information:
\begin{equation}
\small 
\bar{F} = \hat{F}_{micro} + \tilde{F}_{\text{evt}}^{\text{macro}} + \tilde{F}_{\text{rgb}}^{\text{macro}} 
\end{equation}
The additive fusion strategy preserves the original spatial-temporal dimensions while emphasizing overlapping patterns across granularities. Compared to concatenation, summation reduces computational complexity and avoids feature redundancy by inherently reinforcing mutually important signals from both modalities.

\begin{figure} 
\centering
\includegraphics[width=1\linewidth]{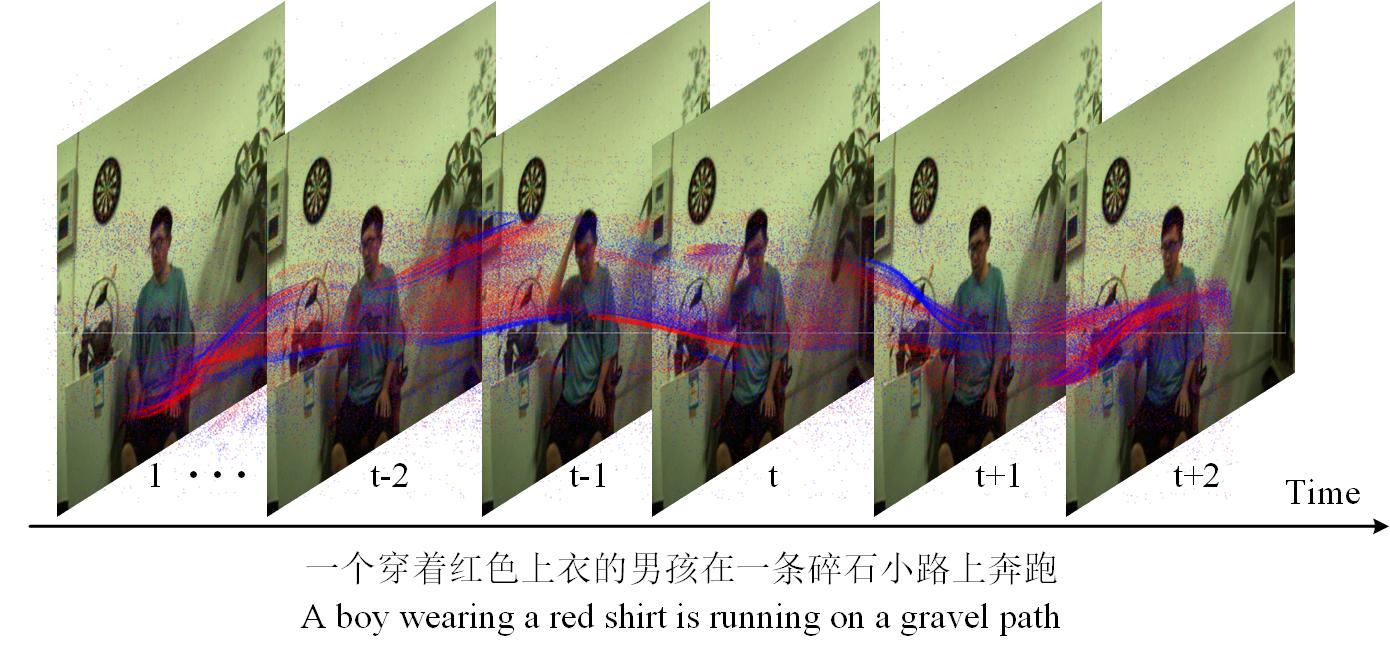}
\caption{A visualization of the RGB frames and Event streams for sign language translation.} 
\label{fig:event_3D_visual}
\end{figure}

In final sign language translation, we use \textit{mBART}~\cite{Liu2020Multilingual} the model pre-trained on CC25~\cite{Liu2020Multilingual} as $\phi_{\text{TextDec}}$. 
The sign language decoder produces a sequence of logits corresponding to the target sentence tokens, formally represented as:
\begin{equation}
\small 
\bar{Y} = \left[\mathbf{z}_0, \mathbf{z}_1, \ldots, \mathbf{z}_T, \mathbf{z}_{T+1}\right]
\end{equation}
where $\mathbf{z}_0$ and $\mathbf{z}_{T+1}$ denote the logits for the beginning of sentence $<$bos$>$ and end-of-sequence $<$eos$>$ tokens respectively, while $\mathbf{z}_t \in \mathbb{R}^{|V|}$ represents the logit scores over vocabulary $V$ at position $t$. Following gloss-free approaches~\cite{zhou2023GFSLT}, we optimize the model using sequence-level cross-entropy loss without intermediate gloss supervision and employ the negative log-likelihood objective that measures the discrepancy between predicted distributions and ground truth labels:
\begin{equation}
\small 
\mathcal{L} = -\sum_{t=1}^{T+1} \log P(s_t | \mathbf{z}_{t-1})
\end{equation}
Here, $s_t$ denotes the target token at timestep $t$, with the summation starting from the first actual token prediction after $<$bos$>$, encourages the model to maximize the likelihood of generating the correct token sequence through autoregressive prediction.


\begin{figure*}
\centering
\includegraphics[width=\linewidth]{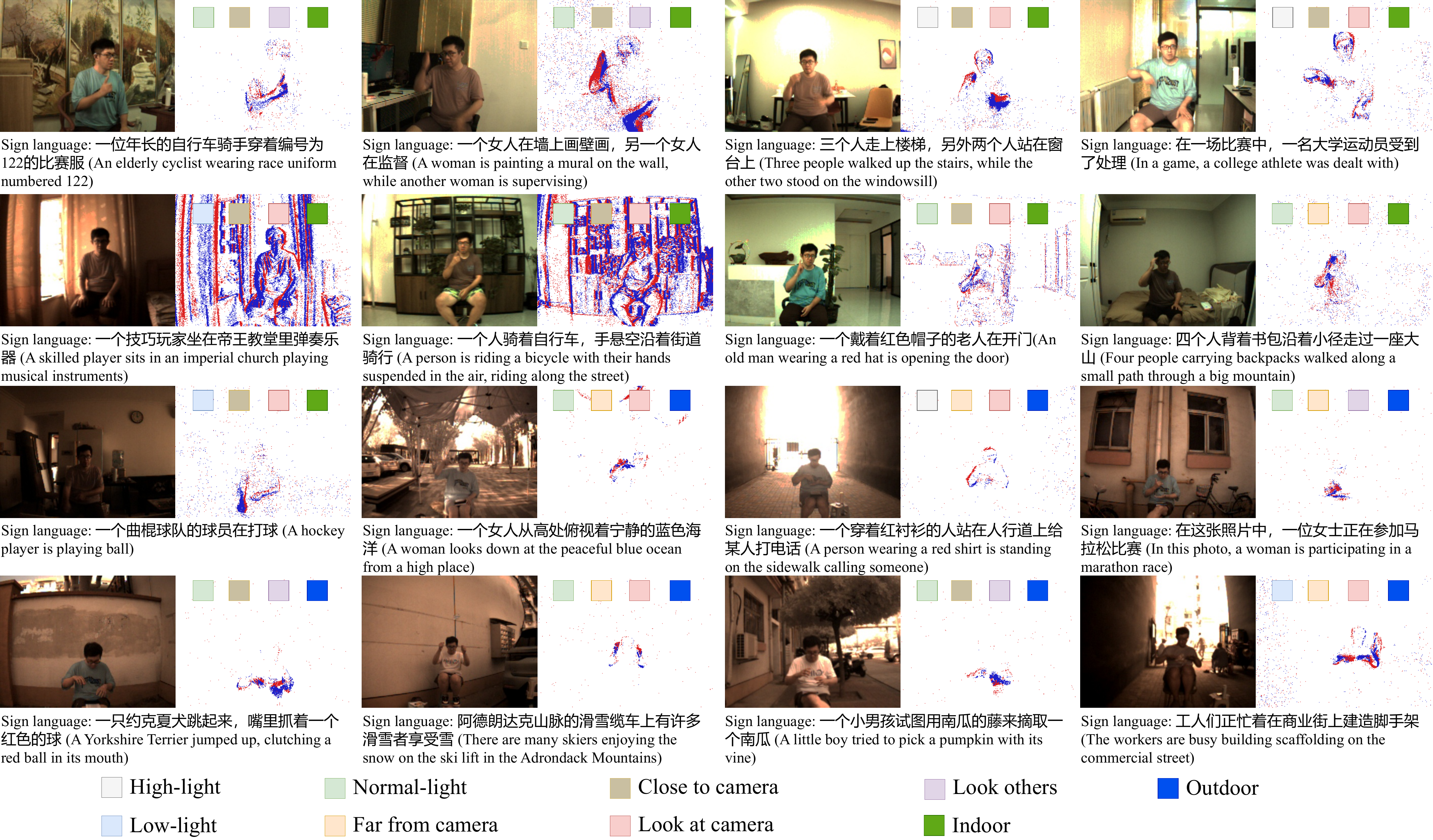}
\caption{Representative samples of our VECSL dataset.}
\label{fig:visualize_dataset}
\end{figure*}

\section{VECSL Benchmark Dataset}

\begin{table*}
\center
\scriptsize  
\caption{Comparison of sign language datasets for sign recognition and translation. The \textit{Continuous} setting refers to scenarios where the sequence of sign language gestures aligns with the sequence of sign vocabulary, as opposed to isolated video clips corresponding to individual glosses.}
\resizebox{\linewidth}{!}{ 
\begin{tabular}{l|llllllcccccc}
\hline \toprule [0.5 pt]
\textbf{Dataset}    &\textbf{Year}  &\textbf{Language}   &\textbf{Videos}   &\textbf{Resolution}  &\textbf{Gloss}  &\textbf{Text}  &\textbf{Indoor}&\textbf{Outdoor}&\textbf{Continuous}  &\textbf{SLT} &\textbf{Event}&\textbf{RGB}\\ 
\hline 
\textbf{SIGNUM}~\cite{von2010signum}            &2010 	&DGS   &$15,075$   &$776\times578$   &$455$   &-&\cmark   &\xmark 
&\cmark 	&\xmark		&\xmark &\cmark	\\ 
\textbf{DEVISIGN-G}~\cite{chai2015devisign}     &2014 	&CSL   &$432$ 	&$640\times480$  &$36$	&-     &\cmark   &\xmark 
&\xmark &\xmark	 &\xmark&\cmark \\
\textbf{DEVISIGN-L}~\cite{chai2015devisign}     &2014		&CSL   &$6,000$ 	&$640\times480$ 	&$500$	&-  	   &\cmark   &\xmark 
&\xmark &\xmark	 &\xmark &\cmark	\\ 
\textbf{DEVISIGN-L}~\cite{chai2015devisign}     &2015 		&CSL   &$24,000$ 		&$640\times480$ 	&$2,000$	&-  	   &\cmark   &\xmark 
&\xmark 		&\xmark	 &\xmark &\cmark	\\ 
\textbf{PHOENIX-2014}~\cite{Koller2015Continuous}  	&2015  	&DGS    &$6,841$ 	&$210\times260$	 &$1,081$ 	&-       &\cmark   &\xmark 
&\cmark 	&\xmark 	&\xmark&\cmark \\
\textbf{PHOENIX-2014T}~\cite{camgoz2018neural}  &2018 	&DGS    &$8,257$  &$210\times260$  	&$1,066$  	 &$2,887$      &\cmark   &\xmark 
&\cmark &\cmark		&\xmark &\cmark \\
\textbf{MS-ASL}~\cite{Joze2018MS-ASL:}&2018 	&ASL     &$25,513$ 		&- 		&$1,000$ 	&-      &\cmark   &\xmark 
&\xmark 		&\xmark		&\xmark&\cmark \\
\textbf{INCLUDE}~\cite{Sridhar2020INCLUDE:}  &2020 	&ISL     &$4,287$ 	&$1920\times1080$ 		&$263$ 	&-  	   &\cmark   &\xmark 
&\xmark 	&\xmark		&\xmark&\cmark \\
\textbf{CSL-Daily}~\cite{zhou2021improving} &2021 	&CSL    &$20,654$   &$1920\times1080$   &$2,000$  &$2,343$      &\cmark   &\xmark 
&\cmark 	&\cmark		&\xmark  &\cmark\\
\hline 
\textbf{SL-Animals-DVS}~\cite{vasudevan2020introduction}  &2020	 &SSL   &$1,121$  &$128\times128$	&$19$ 	&-      &\cmark   &\xmark 
&\xmark	&\xmark		&\cmark	 &\xmark	\\ 
\textbf{EV-ASL}~\cite{Wang2021Event-Based}  &2021	&ASL    &$11,200$ 		&$128\times128$		&$56$ &-  	    &\cmark   &\xmark 
&\xmark 		&\xmark &\cmark	&\xmark  	\\ 
\textbf{EvCSLR}~\cite{jiang2024evcslr} &2024 &CSL   &$2,685$ 	&$346\times260$		&$423$ 	&-     &\cmark   &\xmark 
&\cmark &\xmark  &\cmark		& \cmark	\\ 
\textbf{EvSign}~\cite{zhang2024evsign} &2024 &CSL   &$6,773$ 	&$640\times480$		&$1,387$ 	&$1,947$     &\cmark   &\xmark 
&\cmark &\cmark  &\cmark		& \xmark	\\ 
\hline 
\textbf{VECSL (Ours)}  &2025  &CSL    &$\textbf{15,676}$  &$\mathbf{346\times260}$	&$\textbf{15,191}$ 	&$\textbf{2,568}$     & \cmark   &\cmark   &\cmark  &\cmark	&\cmark &\cmark \\
\hline \toprule [0.5 pt]
\end{tabular}
} 
\label{datasetlist} 
\end{table*}

\subsection{Criteria for Collection and Annotation} 
As illustrated in Fig.~\ref{fig:event_3D_visual} and Fig.~\ref{fig:visualize_dataset}, our \textbf{VECSL} (Vision-Event Chinese Sign Language) dataset addresses conventional sign language datasets' limitations through real-world applicability and multi-modal diversity. The collection and annotation protocols were designed with the following key dimensions:

\noindent $\bullet$ \textbf{Viewpoint Diversity.} Viewpoint diversity enhances dataset utility by capturing signers from multiple angles, simulating natural observation conditions. Our multi-perspective approach overcomes the viewpoint bias typical of fixed-camera laboratory setups.


\noindent $\bullet$ \textbf{Event Camera Motion.} Our dataset includes deliberate camera movements while keeping human subjects stationary. Background noise from camera rotation obscures static silhouettes, and motion blur in RGB frames complicates temporal alignment with event streams. These features make our dataset a robust testbed for dynamic vision systems in real-world, observer-moving scenarios. 

\noindent $\bullet$ \textbf{Scene Complexity.} 
Scene complexity captures real-world usage by incorporating recordings from diverse environments, including indoor spaces (office, living room, corridor) and outdoor locations (public square, sidewalk). The scenarios may introduce dynamic background interference, such as moving vehicles and non-signing individuals, to assess model robustness under realistic conditions.


\noindent $\bullet$ \textbf{Intensity of light.} 
We utilize the high dynamic range of event cameras by capturing data under extreme lighting conditions: high-light scenarios with direct illumination that saturates RGB sensors, normal indoor lighting, and low-light environments with dim illumination that push the limits of conventional camera performance.


\noindent $\bullet$ \textbf{Linguistic Coverage.} The sign language content was sourced from 1,526 COCO-CN~\cite{tmm2019-cococn} captions spanning six categories: daily routines, outdoor activities, animal/plant descriptions, sports commentary, personal narratives, and landmark discussions. Video recordings were guided by these captions to ensure semantic alignment.

\subsection{Statistical Analysis} 

The aforementioned protocol yields 37.52 hours of RGB-Event data (total 15,676 files are stored in \textit{`*.aedat4'} format), establishing our VECSL dataset is well-suited for supporting SLT using single-modal RGB or event data, while also laying a solid foundation for research on multi-modal fusion approaches. All videos in VECSL were captured using Dynamic Vision Sensors (DVS346) cameras with native $346\times260$ resolution. Our dataset comprises 15,676 sign language videos containing 15,191 unique glosses and 2,568 annotated Chinese characters. The dataset is partitioned into training, validation, and testing subsets with 14108, 744, and 824 samples, respectively. 

\begin{table*}[!htp]
\centering 
\caption{Experiment on VECSL dataset. \textsuperscript{$\dagger$} denotes our reproduced result under the gloss-free SLT setting.}
\label{tab:event_metric}
\resizebox{0.9\textwidth}{!}{
\begin{tabular}{lcccccccccccc} 
\hline \toprule [0.5 pt]
\multicolumn{1}{c|}{\textbf{}} & \multicolumn{5}{c|}{\textbf{DEV}} & \multicolumn{5}{c}{\textbf{Test}} \\ \hline
\multicolumn{1}{c|}{\textbf{Method}} & \textbf{BLEU-1} & \textbf{BLEU-2} & \textbf{BLEU-3} & \textbf{BLEU-4}  & \multicolumn{1}{c|}{\textbf{ROUGE}} & \textbf{BLEU-1} & \textbf{BLEU-2} & \textbf{BLEU-3} & \textbf{BLEU-4}  & \textbf{ROUGE} \\ 
\hline \toprule [0.5 pt]
\multicolumn{1}{l|}{TSPNet\textsuperscript{$\dagger$}~\cite{li2020tspnet}     } & 17.97& 10.68& 5.47& 3.24&  \multicolumn{1}{c|}{17.44} & 17.63& 10.66& 5.61& 3.49&  17.34\\
\multicolumn{1}{l|}{Joint-SLT\textsuperscript{$\dagger$}~\cite{camgoz2020sign}     } & 28.38 & 17.65 & 11.09 & 7.53 & \multicolumn{1}{c|}{25.52} & 26.99 & 16.20 & 9.64 & 6.09 & 24.49 \\
\multicolumn{1}{l|}{Sign-XmDA\textsuperscript{$\dagger$}~\cite{ye2023cross}     } & 35.34 & 24.46 & 17.90 & 13.85 &  \multicolumn{1}{c|}{33.26} & 34.42 & 23.72 & 17.11 & 13.04 &  32.02 \\
\multicolumn{1}{l|}{GASLT~\cite{yin2023gloss}    } & 37.72 & 26.98 & 20.12 & 15.81 &  \multicolumn{1}{c|}{34.8} & 37.16 & 26.03 & 18.99 & 14.62  & 33.94 \\
\multicolumn{1}{l|}{GFSLT~\cite{zhou2023GFSLT}    } & 61.73& 52.50& 45.18& 39.54& \multicolumn{1}{c|}{58.67} & 61.93& 52.86& 45.63& 40.01&  58.83\\
\multicolumn{1}{l|}{M$^2$-SLT (Ours)} &69.95&61.43&54.72&49.23&  \multicolumn{1}{c|}{68.92} &70.14&62.03&55.53&50.16& 69.46  \\ 
\hline \toprule [0.5 pt]
\end{tabular}
}
\end{table*}

As demonstrated in Table~\ref{datasetlist}, VECSL uniquely combines three critical characteristics absent in existing sign language datasets: (1) Complex real-world scenes with varied lighting conditions and backgrounds, (2) Continuous sign language sequences rather than isolated words, and (3) Simultaneous event-stream and RGB data capture. This multi-modal representation enables novel research directions in sign language understanding that leverage both frame-based and event-based visual modalities.


\subsection{Benchmark Baselines}  
To establish a reproducible evaluation protocol for gloss-free sign language translation in RGB-Event bimodal settings, we reconfigure five state-of-the-art (SOTA) methods spanning TSPNet~\cite{li2020tspnet}, Joint-SLT~\cite{camgoz2020sign}, Sign-XmDA~\cite{ye2023cross}, GASLT~\cite{yin2023gloss}, GFSLT~\cite{zhou2023GFSLT}.
All baselines are adapted through two essential modifications: 
1) Following existing works~\cite{wang2024visevent}, event frames and RGB frames are processed through a weighted linear combination for unimodal frameworks, i.e., $X_{\text{fused}} = 0.5 \cdot X_{\text{rgb}} + 0.5 \cdot X_{\text{evt}}$, preserving original architecture compatibility while enabling bimodal awareness; 
2) Gloss dependency elimination via direct Chinese character substitution, enabling end-to-end translation without intermediate linguistic representations.

\section{Experiment} 

\subsection{Dataset and Evaluation Metric}  
In the absence of any publicly available RGB-Event SLT datasets, the experiments were conducted and validated on our newly proposed VECSL dataset. 
We adopt BLEU~\cite{papineni2002bleu} for n-gram precision and ROUGE-L~\cite{lin2004rouge} for content recall as metrics. BLEU evaluates lexical alignment via smoothed multi-reference n-gram matching, while ROUGE-L calculates the F-measure based on the longest common subsequence, emphasizing content preservation over exact phrasing.
Higher scores on both indicate improved translation quality, with their combination offering a balanced assessment.

\subsection{Implementation Details} 
\label{sec:implementation_details}

For the MiR module, we initialize the micro-sign with 128 learnable features ($\hat{L}=128$), each with a dimension of 512. The top-k retrieval is set to k=3. For the MaR module, we utilize Video Mamba~\cite{li2024videomamba} to extract video features from the training set, with the macro-sign features $\tilde{F}_{\text{*}}^{\text{macro}}$, resulting in 123 distinct features. We optimize the model using SGD optimizer~\cite{bottou2012stochastic} with an initial learning rate of 0.01 and a cosine annealing scheduler for learning rate adjustment. To accelerate training, we perform uniform frame sampling with an interval of 4 and a maximum input frame count of 64. All experiments are conducted on a server equipped with NVIDIA A800 80GB GPUs. More details can be found in our source code.

\subsection{Comparison on VECSL} 

As shown in Table~\ref{tab:event_metric}, we provide a comprehensive comparison between our method and several SOTA approaches for sign language translation on the VECSL dataset. 
Our approach achieves superior performance, with ROUGE and BLEU-4 scores of 69.46 and 50.16, respectively. In contrast, previous methods such as TSPNet~\cite{li2020tspnet}, Joint-SLT~\cite{camgoz2020sign}, Sign-XmDA~\cite{ye2023cross}, and GASLT~\cite{yin2023gloss} exhibit relatively limited performance, which may be attributed to their reliance on frozen encoders for feature extraction during training. Notably, compared to GFSLT~\cite{yin2023gloss}, which also operates without a pre-training stage, our method achieves improvements of +10.63 in ROUGE and +10.15 in BLEU-4. 

\subsection{Ablation Study}   




\noindent $\bullet$ \textbf{Necessity of RGB-Event Fusion Strategies for SLT.}  %
In this subsection, we explore the necessity of fusion strategies for combining RGB and event data in SLT. We conducted experiments using only the visual encoder and text decoder to isolate the impact of fusion strategies. The experimental results in Table~\ref{easy_fusion_strategies} reveal significant performance degradation in naive fusion approaches (Rows \#02-\#04). Specifically, Row 01 represents the baseline using only RGB input. Row 02 employs a weighted fusion by \(X_{\text{fused}} = 0.5 \cdot X_{\text{rgb}} + 0.5 \cdot X_{\text{evt}}\). Row 03 processes RGB and event data separately and concatenates their features before feeding them into the text decoder. Row 04 sums the features of RGB and event data. The observed degradation in these naive fusion approaches suggests that while event streams inherently capture high-temporal-frequency micro-articulatory trajectories unavailable in RGB, simple fusion methods fail to effectively integrate this information. In contrast, our proposed method (Row 05) improves the BLEU-4 score by +1.58 compared to RGB-Event Feature Sum (Row 04), demonstrating the effectiveness of our retrieval-augmented fusion strategy.

\begin{table}
\center
\caption{Experimental study on RGB-Event fusion strategies for our framework. \textsuperscript{\textdaggerdbl} denote utilizing only the visual and language modules for video encoding and sign language decoding, without incorporating MiR and MaR modules}
\label{easy_fusion_strategies}
\resizebox{\linewidth}{!}{
\begin{tabular}{c|l|c|c|c|c|c}
\hline \toprule [0.5 pt]
\textbf{No.} &\textbf{Method}  &\textbf{BLEU-1} &\textbf{BLEU-2} &\textbf{BLEU-3} &\textbf{BLEU-4} &\textbf{ROUGE} \\
\hline
01 &RGB-only\textsuperscript{\textdaggerdbl}  &70.29&61.81&54.91&49.25&69.00 \\ 
02 &Frames Weighted\textsuperscript{\textdaggerdbl} &68.32&59.77&52.94&47.40&66.53\\ 
03 &Feature Concat\textsuperscript{\textdaggerdbl} &69.32&61.0&54.34&48.84&68.66 \\ 
04 &Feature Sum\textsuperscript{\textdaggerdbl}  &68.33&60.28&53.82&48.58&67.99 \\ 
05 &Ours  & 70.14&62.03&55.53&50.16&69.46 \\ 
\hline \toprule [0.5 pt] 
\end{tabular}
}
\end{table}

\noindent $\bullet$ \textbf{Ablation of MiR and MaR.} 
As shown in Table~\ref{tab:ablation_MiR_MaR}, we conduct an ablation study on the proposed MiR, MaR, and visual encoder. Comparing Rows 01 and 02, replacing ResNet18 with SignGraph as the visual encoder improves BLEU-4 by $+0.50$, despite a slight decrease in BLEU-1 ($-0.70$). The improvement suggests SignGraph better captures fine-grained temporal dependencies critical for higher-order n-gram accuracy. Enabling MiR's micro-sign retrieval (Row 04 vs. Row 02) boosts BLEU-4 by $+0.66$ , while adding recurrent processing in MiR (Row 05 vs. Row 04) further improves BLEU-4 by $+0.08$. Similarly, activating MaR (Row 03 vs. Row 02) enhances BLEU-4 by $+0.62$, demonstrating its effectiveness in modeling macro-sign transitions. When all modules are combined (Row 06 vs. Row 01), we achieve a $+2.08$ BLEU-4 improvement (48.08 → 50.16), validating the complementary roles of MiR and MaR in disentangling micro- and macro-sign semantics.

\begin{table}
\center
\small    
\caption{Ablation study of the MiR and MaR modules. Row 01 replaces SignGraph with ResNet18, while the full model in Row 06 achieves the best results.}
\label{tab:ablation_MiR_MaR}
\resizebox{\linewidth}{!}{
\begin{tabular}{c|c|cc|c|c|c|c|c}
\hline \toprule [0.5 pt]
\multirow{2}{*}{\textbf{No.}} &\multirow{2}{*}{\textbf{SignGraph}} & \multicolumn{2}{c|}{\textbf{MiR}} & \multirow{2}{*}{\textbf{MaR}} & \multirow{2}{*}{\textbf{B-1}} & \multirow{2}{*}{\textbf{B-2}} & \multirow{2}{*}{\textbf{B-3}} & \multirow{2}{*}{\textbf{B-4}} \\ \cline{3-4}
 & & \textbf{Micro Sign} & \textbf{Recurrent} &  &  &  &  &  \\ \hline
01 &\xmark& \xmark & \xmark & \xmark &69.03&60.48&53.62&48.08 \\
02 &\cmark& \xmark & \xmark & \xmark & 68.33 & 60.28 & 53.82 & 48.58 \\
03 &\cmark& \xmark & \xmark & \cmark & 69.73 & 61.39 & 54.65 & 49.20 \\
04 &\cmark& \cmark & \xmark & \xmark & 69.62 & 61.50 & 54.75 & 49.24 \\
05 &\cmark& \cmark & \cmark & \xmark & 70.05 & 61.66 & 54.84 & 49.32 \\
06 &\cmark& \cmark & \cmark & \cmark & 70.14 & 62.03 & 55.53 & 50.16 \\ 
\hline \toprule [0.5 pt] 
\end{tabular}
}
\end{table}

\noindent $\bullet$ \textbf{Parameter analysis of Micro-Sign.} 
As shown in Table~\ref{tab:Micro_sign_Parameter}, retrieving the Top-60 most relevant micro-sign features yields a 0.75 BLEU-4 score reduction compared to using the Top-3 features, which performance degradation suggests that excessive feature retrieval introduces either noise or redundant temporal patterns, potentially overwhelming the model's capacity to prioritize discriminative micro-level cues. Moreover, increasing the micro-sign memory feature dimension from 512 to 2048 results in a more substantial performance drop of 4.56 BLEU-4 points. The underlying reasons could be projection-layer inefficiency when mapping 1024 inputs to 2048 (vs. 512), where high-dimensional embeddings may dilute critical spatiotemporal relationships

\begin{table}
\center
\caption{Micro-sign parameters setting. 'Dim' denote micro-sign memory feature dimension.}
\label{tab:Micro_sign_Parameter}
\resizebox{\linewidth}{!}{
\begin{tabular}{c|l|c|c|c|c|c}
\hline \toprule [0.5 pt]
\textbf{No.} &\textbf{Method}  &\textbf{BLEU-1} &\textbf{BLEU-2} &\textbf{BLEU-3} &\textbf{BLEU-4} &\textbf{ROUGE} \\
\hline
01 & Top-64   & 69.77&61.56&54.87&49.41&69.21 \\ 
02 & Dim-2048  &67.51&58.56&51.36&45.60&66.53\\ 
03 & Top-3 + Dim-512& 70.14&62.03&55.53&50.16&69.46 \\ 
\hline \toprule [0.5 pt] 
\end{tabular}
}
\end{table}

\noindent $\bullet$ \textbf{Parameter analysis of Macro-Sign.} 
We conduct a comprehensive evaluation of macro-sign encoding methods to assess their impact on translation performance. As shown in Table~\ref{tab:Macro_sign_Parameter}, X3D~\cite{feichtenhofer2020x3d} achieves the highest scores across all metrics, with a BLEU-4 score of 51.28, outperforming I3D~\cite{carreira2017I3D}, Slowfast~\cite{feichtenhofer2019slowfast}, and Video Mamba~\cite{li2024videomamba}. Notably, X3D’s superior performance suggests its spatiotemporal feature extraction better captures macro-level sign language patterns, particularly in modeling long-range dependencies and hierarchical temporal structures. Although Video Mamba achieves a slightly lower BLEU-4 score of 50.16, its competitive performance and unique ability to temporal dynamics make it a suitable choice for our framework.

\begin{table}
\center
\caption{Selection of Video encoder on VECSL dataset.}
\label{tab:Macro_sign_Parameter}
\resizebox{\linewidth}{!}{
\begin{tabular}{c|l|c|c|c|c|c}
\hline \toprule [0.5 pt]
\textbf{No.} &\textbf{Method}   &\textbf{BLEU-1} &\textbf{BLEU-2} &\textbf{BLEU-3} &\textbf{BLEU-4} &\textbf{ROUGE} \\
\hline
01 &I3D~\cite{carreira2017I3D}   &70.01&61.73&55.09&49.70&68.99 \\ 
02 &X3D~\cite{feichtenhofer2020x3d}    &72.29&63.89&56.94&51.28&71.22 \\ 
03 &Slowfast~\cite{feichtenhofer2019slowfast}    & 	70.29&62.14&55.51&50.09&69.09 \\ 
04 &Video Mamba~\cite{li2024videomamba}  & 70.14&62.03&55.53&50.16&69.46 \\ 
\hline \toprule [0.5 pt] 
\end{tabular}
}
\end{table}

\noindent $\bullet$ \textbf{Ablation Study on Feature Fusion Strategies between MaR and MiR.}
We analyze the impact of different feature fusion strategies for integrating macro-sign and micro-sign features before the text decoder. As shown in Table~\ref{tab:MaR_MiR_fusion}, Feature Sum significantly outperforms the Feature Concat approach across all metrics, achieving a BLEU-4 score of 50.16 compared to 42.50 for feature concatenate before text decoder. Preserving feature space dimensionality, the sum operation avoids increased complexity introduced by concatenation, which may dilute the model’s capacity to effectively leverage both macro and micro-level cues. Consequently, suboptimal translation results arise from concatenation’s inherent limitations.

\begin{table}
\center
\caption{Analysis of the feature fusion strategies of MiR and MaR before text decoder.}
\label{tab:MaR_MiR_fusion}
\small    
\resizebox{\linewidth}{!}{
\begin{tabular}{c|l|c|c|c|c|c}
\hline \toprule [0.5 pt]
\textbf{No.} &\textbf{Method}   &\textbf{BLEU-1} &\textbf{BLEU-2} &\textbf{BLEU-3} &\textbf{BLEU-4} &\textbf{ROUGE}\\
\hline
01 &Feature Concat   &65.20&55.83&48.37&42.50&62.93\\ 
02 &Feature Sum  &70.14&62.03&55.53&50.16&69.46 \\ 
\hline 
\toprule [0.5 pt]
\end{tabular}
}
\end{table}


\begin{figure}[!htp]
\centering
\small
\includegraphics[width=1\linewidth]{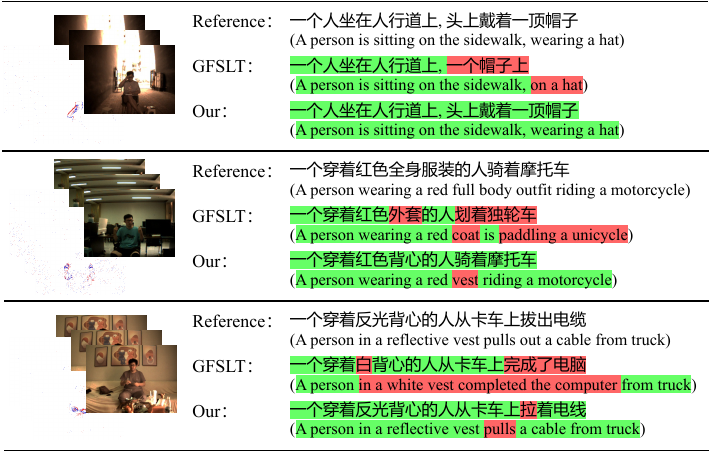}
\caption{Qualitative results of VECSL. Semantically incorrect sentences generated by the model are highlighted in red, while semantically correct sentences are highlighted in green.}
\label{fig:Qualitative_results}
\end{figure}

\subsection{Visualization} \label{sec:Visualization}


\noindent $\bullet$ \textbf{Qualitative Comparison.} 
As shown in Figure ~\ref{fig:Qualitative_results}, we present several examples to validate the effectiveness of our proposed model. For a given RGB-Event video, we compare the \textit{Reference} with the outputs generated by both our model and the GFSLT~\cite{zhou2023GFSLT} model to ensure a more comprehensive and objective evaluation. To enhance visual clarity, we highlight the generated results that align with the \textit{Reference} in different colors: green highlights indicate segments where both models' outputs match the \textit{Reference} in semantics, while red highlights mark discrepancies. As evident from the figure, our model produces results that are more closely aligned with the \textit{Reference} and demonstrate higher accuracy compared to the GFSLT~\cite{zhou2023GFSLT} model.

\begin{figure}[!htp]
\centering
\small
\includegraphics[width=1\linewidth]{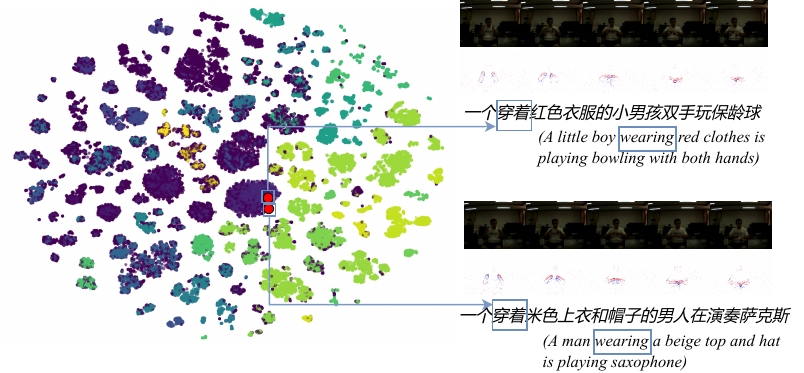}
\caption{T-SNE visualization of features from two distinct videos, where action proposals corresponding to "wearing" are closely clustered in the t-SNE space.}
\label{fig:MiR_tsne}
\end{figure}

\noindent $\bullet$ \textbf{T-SNE Visualization of Video Encoded Features in Macro-Sign.}
As illustrated in Fig.~\ref{fig:MiR_tsne}, features encoded by the video encoder are reduced to 2D using t-SNE. Observations reveal that visual features corresponding to the same text exhibit close proximity in the t-SNE space. For instance, sign language actions representing the word "wearing" form tight clusters. The geometric alignment proximity between feature points demonstrates the encoder's capability to extract semantic similarities across diverse video instances, which facilitates macro-sign-based retrieval.

\subsection{Limitation Analysis}  
While our framework demonstrates superior performance over existing models, there remains room for improvement.  The primary limitation stems from the computational constraints addressed through the adoption of a lightweight decoder in the sign language generation stage, where integrating large-scale language models could substantially enhance performance.  Additionally, the framework’s reliance on extensive paired high-quality training data partially restricts its generalization capabilities across diverse contextual scenarios, a common challenge shared by current sign language recognition frameworks.  

\section{Conclusion}  

In this paper, we present the first large-scale RGB-Event sign language dataset, VECSL, which not only boasts significant advantages in terms of scale and diversity but also offers a novel perspective for sign language translation through the integration of event data.  In parallel, we propose a sign language translation framework that fuses information from both RGB and event streams.  The framework incorporates MiR and MaR modules to enhance the capture of local fine-grained details and global semantic information, while cross-modal feature alignment further improves the model's robustness and generalization.  Experimental results demonstrate that our framework significantly enhances the capture of microscopic motion details and overall semantic expression, outperforming existing approaches.  
In future work, we will focus on leveraging larger decoders to achieve superior performance without increasing computational complexity, and explore unsupervised or weakly supervised learning methods to reduce reliance on large-scale annotated data.

{
    \small
    \bibliographystyle{ieeenat_fullname}
    \bibliography{reference}

\begin{thebibliography}{62}
\providecommand{\natexlab}[1]{#1}
\providecommand{\url}[1]{\texttt{#1}}
\expandafter\ifx\csname urlstyle\endcsname\relax
  \providecommand{\doi}[1]{doi: #1}\else
  \providecommand{\doi}{doi: \begingroup \urlstyle{rm}\Url}\fi

\bibitem[Blattmann et~al.(2022)Blattmann, Rombach, Oktay, M{\"u}ller, and
  Ommer]{blattmann2022retrieval}
Andreas Blattmann, Robin Rombach, Kaan Oktay, Jonas M{\"u}ller, and Bj{\"o}rn
  Ommer.
\newblock Retrieval-augmented diffusion models.
\newblock \emph{Advances in Neural Information Processing Systems},
  35:\penalty0 15309--15324, 2022.

\bibitem[Bottou(2012)]{bottou2012stochastic}
L{\'e}on Bottou.
\newblock Stochastic gradient descent tricks.
\newblock In \emph{Neural Networks: Tricks of the Trade: Second Edition}, pages
  421--436. Springer, 2012.

\bibitem[Camgoz et~al.(2018)Camgoz, Hadfield, Koller, Ney, and
  Bowden]{camgoz2018neural}
Necati~Cihan Camgoz, Simon Hadfield, Oscar Koller, Hermann Ney, and Richard
  Bowden.
\newblock Neural sign language translation.
\newblock In \emph{Proceedings of the IEEE conference on computer vision and
  pattern recognition}, pages 7784--7793, 2018.

\bibitem[Camgoz et~al.(2020{\natexlab{a}})Camgoz, Koller, Hadfield, and
  Bowden]{camgoz2020multi}
Necati~Cihan Camgoz, Oscar Koller, Simon Hadfield, and Richard Bowden.
\newblock Multi-channel transformers for multi-articulatory sign language
  translation.
\newblock In \emph{Computer Vision--ECCV 2020 Workshops: Glasgow, UK, August
  23--28, 2020, Proceedings, Part IV 16}, pages 301--319. Springer,
  2020{\natexlab{a}}.

\bibitem[Camgoz et~al.(2020{\natexlab{b}})Camgoz, Koller, Hadfield, and
  Bowden]{camgoz2020sign}
Necati~Cihan Camgoz, Oscar Koller, Simon Hadfield, and Richard Bowden.
\newblock Sign language transformers: Joint end-to-end sign language
  recognition and translation.
\newblock In \emph{Proceedings of the IEEE/CVF conference on computer vision
  and pattern recognition}, pages 10023--10033, 2020{\natexlab{b}}.

\bibitem[Carreira and Zisserman(2017)]{carreira2017I3D}
Joao Carreira and Andrew Zisserman.
\newblock Quo vadis, action recognition? a new model and the kinetics dataset.
\newblock In \emph{proceedings of the IEEE Conference on Computer Vision and
  Pattern Recognition}, pages 6299--6308, 2017.

\bibitem[Chai et~al.(2015)Chai, Wanga, Zhoub, Wub, Lic, and
  Chena]{chai2015devisign}
X Chai, H Wanga, M Zhoub, G Wub, H Lic, and X Chena.
\newblock Devisign: dataset and evaluation for 3d sign language recognition.
\newblock \emph{Technical report, Beijing, Tech. Rep}, 2015.

\bibitem[Chaudhary et~al.(2023)Chaudhary, Ananthanarayana, Hoq, and
  Nwogu]{Chaudhary2023SignNet}
Lipisha Chaudhary, Tejaswini Ananthanarayana, Enjamamul Hoq, and Ifeoma Nwogu.
\newblock Signnet ii: A transformer-based two-way sign language translation
  model.
\newblock \emph{IEEE Transactions on Pattern Analysis and Machine
  Intelligence}, 45\penalty0 (11), 2023.

\bibitem[Chen et~al.(2022)Chen, Wei, Sun, Wu, and Lin]{chen2022simple}
Yutong Chen, Fangyun Wei, Xiao Sun, Zhirong Wu, and Stephen Lin.
\newblock A simple multi-modality transfer learning baseline for sign language
  translation.
\newblock In \emph{Proceedings of the IEEE/CVF conference on computer vision
  and pattern recognition}, pages 5120--5130, 2022.

\bibitem[Cheng et~al.(2023)Cheng, Luo, Chen, Liu, Zhao, and Yan]{cheng2023lift}
Xin Cheng, Di Luo, Xiuying Chen, Lemao Liu, Dongyan Zhao, and Rui Yan.
\newblock Lift yourself up: Retrieval-augmented text generation with
  self-memory.
\newblock \emph{Advances in Neural Information Processing Systems},
  36:\penalty0 43780--43799, 2023.

\bibitem[Cheng et~al.(2025)Cheng, Wang, Zhang, Ge, Chen, Wei, Zhang, and
  Zhao]{cheng2025xrag}
Xin Cheng, Xun Wang, Xingxing Zhang, Tao Ge, Si-Qing Chen, Furu Wei, Huishuai
  Zhang, and Dongyan Zhao.
\newblock xrag: Extreme context compression for retrieval-augmented generation
  with one token.
\newblock \emph{Advances in Neural Information Processing Systems},
  37:\penalty0 109487--109516, 2025.

\bibitem[Ester et~al.(1996{\natexlab{a}})Ester, Kriegel, Sander, Xu,
  et~al.]{ester1996dbscan}
Martin Ester, Hans-Peter Kriegel, J{\"o}rg Sander, Xiaowei Xu, et~al.
\newblock A density-based algorithm for discovering clusters in large spatial
  databases with noise.
\newblock In \emph{kdd}, pages 226--231, 1996{\natexlab{a}}.

\bibitem[Ester et~al.(1996{\natexlab{b}})Ester, Kriegel, Sander, Xu,
  et~al.]{ester1996density}
Martin Ester, Hans-Peter Kriegel, J{\"o}rg Sander, Xiaowei Xu, et~al.
\newblock A density-based algorithm for discovering clusters in large spatial
  databases with noise.
\newblock In \emph{kdd}, pages 226--231, 1996{\natexlab{b}}.

\bibitem[Feichtenhofer(2020)]{feichtenhofer2020x3d}
Christoph Feichtenhofer.
\newblock X3d: Expanding architectures for efficient video recognition.
\newblock In \emph{Proceedings of the IEEE/CVF conference on computer vision
  and pattern recognition}, pages 203--213, 2020.

\bibitem[Feichtenhofer et~al.(2019)Feichtenhofer, Fan, Malik, and
  He]{feichtenhofer2019slowfast}
Christoph Feichtenhofer, Haoqi Fan, Jitendra Malik, and Kaiming He.
\newblock Slowfast networks for video recognition.
\newblock In \emph{Proceedings of the IEEE/CVF international conference on
  computer vision}, pages 6202--6211, 2019.

\bibitem[Feng et~al.(2025)Feng, Yu, Cheng, Tang, Zhang, Yuan, and
  Tian]{feng2025aeNeRF}
Chaoran Feng, Wangbo Yu, Xinhua Cheng, Zhenyu Tang, Junwu Zhang, Li Yuan, and
  Yonghong Tian.
\newblock Ae-nerf: Augmenting event-based neural radiance fields for non-ideal
  conditions and larger scene.
\newblock \emph{arXiv preprint arXiv:2501.02807}, 2025.

\bibitem[Gan et~al.(2024)Gan, Yin, Jiang, Wen, Xie, and Lu]{gan2024signgraph}
Shiwei Gan, Yafeng Yin, Zhiwei Jiang, Hongkai Wen, Lei Xie, and Sanglu Lu.
\newblock Signgraph: A sign sequence is worth graphs of nodes.
\newblock In \emph{Proceedings of the IEEE/CVF Conference on Computer Vision
  and Pattern Recognition}, pages 13470--13479, 2024.

\bibitem[Gehrig and Scaramuzza(2023)]{gehrig2023rvt}
Mathias Gehrig and Davide Scaramuzza.
\newblock Recurrent vision transformers for object detection with event
  cameras.
\newblock In \emph{Proceedings of the IEEE/CVF conference on computer vision
  and pattern recognition}, pages 13884--13893, 2023.

\bibitem[Gong et~al.(2024)Gong, Foo, He, Rahmani, and Liu]{gong2024llms}
Jia Gong, Lin~Geng Foo, Yixuan He, Hossein Rahmani, and Jun Liu.
\newblock Llms are good sign language translators.
\newblock In \emph{Proceedings of the IEEE/CVF Conference on Computer Vision
  and Pattern Recognition}, pages 18362--18372, 2024.

\bibitem[Guo et~al.(2018)Guo, Zhou, Li, and Wang]{guo2018hierarchical}
Dan Guo, Wengang Zhou, Houqiang Li, and Meng Wang.
\newblock Hierarchical lstm for sign language translation.
\newblock In \emph{Proceedings of the AAAI conference on artificial
  intelligence}, 2018.

\bibitem[Hamann et~al.(2024)Hamann, Ghosh, Martinez, Hart, Kacelnik, and
  Gallego]{hamann2024reTAG}
Friedhelm Hamann, Suman Ghosh, Ignacio~Juarez Martinez, Tom Hart, Alex
  Kacelnik, and Guillermo Gallego.
\newblock Low-power continuous remote behavioral localization with event
  cameras.
\newblock In \emph{Proceedings of the IEEE/CVF Conference on Computer Vision
  and Pattern Recognition}, pages 18612--18621, 2024.

\bibitem[He et~al.(2016)He, Zhang, Ren, and Sun]{he2016resnet}
Kaiming He, Xiangyu Zhang, Shaoqing Ren, and Jian Sun.
\newblock Deep residual learning for image recognition.
\newblock In \emph{Proceedings of the IEEE conference on computer vision and
  pattern recognition}, pages 770--778, 2016.

\bibitem[Jiang et~al.(2024)Jiang, Wang, Li, Zhang, Guo, Chu, and
  Gao]{jiang2024evcslr}
Yu Jiang, Yuehang Wang, Siqi Li, Yongji Zhang, Qianren Guo, Qi Chu, and Yue
  Gao.
\newblock Evcslr: Event-guided continuous sign language recognition and
  benchmark.
\newblock \emph{IEEE Transactions on Multimedia}, 2024.

\bibitem[Jiang et~al.(2023)Jiang, Xu, Gao, Sun, Liu, Dwivedi-Yu, Yang, Callan,
  and Neubig]{jiang2023activeRAG}
Zhengbao Jiang, Frank~F Xu, Luyu Gao, Zhiqing Sun, Qian Liu, Jane Dwivedi-Yu,
  Yiming Yang, Jamie Callan, and Graham Neubig.
\newblock Active retrieval augmented generation.
\newblock In \emph{Proceedings of the 2023 Conference on Empirical Methods in
  Natural Language Processing}, pages 7969--7992, 2023.

\bibitem[Joze and Koller(2018)]{Joze2018MS-ASL:}
Hamid Reza~Vaezi Joze and Oscar Koller.
\newblock Ms-asl: A large-scale data set and benchmark for understanding
  american sign language.
\newblock In \emph{British Machine Vision Conference}, page 100, 2018.

\bibitem[Koller et~al.(2015)Koller, Forster, and Ney]{Koller2015Continuous}
Oscar Koller, Jens Forster, and Hermann Ney.
\newblock Continuous sign language recognition: Towards large vocabulary
  statistical recognition systems handling multiple signers.
\newblock \emph{Computer Vision and Image Understanding}, 141\penalty0
  (1):\penalty0 108--125, 2015.

\bibitem[Lewis et~al.(2020)Lewis, Perez, Piktus, Petroni, Karpukhin, Goyal,
  K{\"u}ttler, Lewis, Yih, Rockt{\"a}schel, et~al.]{lewis2020retrieval}
Patrick Lewis, Ethan Perez, Aleksandra Piktus, Fabio Petroni, Vladimir
  Karpukhin, Naman Goyal, Heinrich K{\"u}ttler, Mike Lewis, Wen-tau Yih, Tim
  Rockt{\"a}schel, et~al.
\newblock Retrieval-augmented generation for knowledge-intensive nlp tasks.
\newblock \emph{Advances in neural information processing systems},
  33:\penalty0 9459--9474, 2020.

\bibitem[Li et~al.(2020)Li, Xu, Yu, Zhang, Swift, Suominen, and
  Li]{li2020tspnet}
Dongxu Li, Chenchen Xu, Xin Yu, Kaihao Zhang, Benjamin Swift, Hanna Suominen,
  and Hongdong Li.
\newblock Tspnet: Hierarchical feature learning via temporal semantic pyramid
  for sign language translation.
\newblock \emph{Advances in Neural Information Processing Systems},
  33:\penalty0 12034--12045, 2020.

\bibitem[Li et~al.(2024)Li, Li, Wang, He, Wang, Wang, and
  Qiao]{li2024videomamba}
Kunchang Li, Xinhao Li, Yi Wang, Yinan He, Yali Wang, Limin Wang, and Yu Qiao.
\newblock Videomamba: State space model for efficient video understanding.
\newblock In \emph{European Conference on Computer Vision}, pages 237--255.
  Springer, 2024.

\bibitem[Li et~al.(2019)Li, Xu, Wang, Lan, Jia, Yang, and Xu]{tmm2019-cococn}
Xirong Li, Chaoxi Xu, Xiaoxu Wang, Weiyu Lan, Zhengxiong Jia, Gang Yang, and
  Jieping Xu.
\newblock Coco-cn for cross-lingual image tagging, captioning and retrieval.
\newblock \emph{IEEE Transactions on Multimedia}, 21\penalty0 (9):\penalty0
  2347--2360, 2019.

\bibitem[Lin(2004)]{lin2004rouge}
Chin-Yew Lin.
\newblock Rouge: A package for automatic evaluation of summaries.
\newblock In \emph{Text summarization branches out}, pages 74--81, 2004.

\bibitem[Liu et~al.(2023)Liu, Son, Yang, Liu, Gao, Lee, and
  Li]{liu2023learning}
Haotian Liu, Kilho Son, Jianwei Yang, Ce Liu, Jianfeng Gao, Yong~Jae Lee, and
  Chunyuan Li.
\newblock Learning customized visual models with retrieval-augmented knowledge.
\newblock In \emph{Proceedings of the IEEE/CVF Conference on Computer Vision
  and Pattern Recognition}, pages 15148--15158, 2023.

\bibitem[Liu et~al.(2020)Liu, Gu, Goyal, Li, Edunov, Ghazvininejad, Lewis, and
  Zettlemoyer]{Liu2020Multilingual}
Yinhan Liu, Jiatao Gu, Naman Goyal, Xian Li, Sergey Edunov, Marjan
  Ghazvininejad, Michael Lewis, and Luke Zettlemoyer.
\newblock Multilingual denoising pre-training for neural machine translation.
\newblock \emph{Transactions of the Association for Computational Linguistics},
  8:\penalty0 726--742, 2020.

\bibitem[Long et~al.(2022)Long, Yin, Ajanthan, Nguyen, Purkait, Garg, Blair,
  Shen, and van~den Hengel]{long2022retrieval}
Alexander Long, Wei Yin, Thalaiyasingam Ajanthan, Vu Nguyen, Pulak Purkait,
  Ravi Garg, Alan Blair, Chunhua Shen, and Anton van~den Hengel.
\newblock Retrieval augmented classification for long-tail visual recognition.
\newblock In \emph{Proceedings of the IEEE/CVF conference on computer vision
  and pattern recognition}, pages 6959--6969, 2022.

\bibitem[Luo et~al.(2024)Luo, Luo, Wang, Lin, Zeng, and Liu]{luo2024EEMFlow}
Xinglong Luo, Ao Luo, Zhengning Wang, Chunyu Lin, Bing Zeng, and Shuaicheng
  Liu.
\newblock Efficient meshflow and optical flow estimation from event cameras.
\newblock In \emph{Proceedings of the IEEE/CVF Conference on Computer Vision
  and Pattern Recognition}, pages 19198--19207, 2024.

\bibitem[Papineni et~al.(2002)Papineni, Roukos, Ward, and
  Zhu]{papineni2002bleu}
Kishore Papineni, Salim Roukos, Todd Ward, and Wei-Jing Zhu.
\newblock Bleu: a method for automatic evaluation of machine translation.
\newblock In \emph{Proceedings of the 40th annual meeting of the Association
  for Computational Linguistics}, pages 311--318, 2002.

\bibitem[Peng et~al.(2023)Peng, Zhang, Xiong, Sun, and Wu]{peng2023get}
Yansong Peng, Yueyi Zhang, Zhiwei Xiong, Xiaoyan Sun, and Feng Wu.
\newblock Get: Group event transformer for event-based vision.
\newblock In \emph{Proceedings of the IEEE/CVF International Conference on
  Computer Vision}, pages 6038--6048, 2023.

\bibitem[Ram et~al.(2023)Ram, Levine, Dalmedigos, Muhlgay, Shashua,
  Leyton-Brown, and Shoham]{ram2023context}
Ori Ram, Yoav Levine, Itay Dalmedigos, Dor Muhlgay, Amnon Shashua, Kevin
  Leyton-Brown, and Yoav Shoham.
\newblock In-context retrieval-augmented language models.
\newblock \emph{Transactions of the Association for Computational Linguistics},
  11:\penalty0 1316--1331, 2023.

\bibitem[Ramsauer et~al.(2020)Ramsauer, Sch{\"a}fl, Lehner, Seidl, Widrich,
  Adler, Gruber, Holzleitner, Pavlovi{\'c}, Sandve,
  et~al.]{ramsauer2020hopfield}
Hubert Ramsauer, Bernhard Sch{\"a}fl, Johannes Lehner, Philipp Seidl, Michael
  Widrich, Thomas Adler, Lukas Gruber, Markus Holzleitner, Milena Pavlovi{\'c},
  Geir~Kjetil Sandve, et~al.
\newblock Hopfield networks is all you need.
\newblock \emph{arXiv preprint arXiv:2008.02217}, 2020.

\bibitem[Sridhar et~al.(2020)Sridhar, Ganesan, Kumar, and
  Khapra]{Sridhar2020INCLUDE:}
Advaith Sridhar, Rohith~Gandhi Ganesan, Pratyush Kumar, and Mitesh Khapra.
\newblock Include: A large scale dataset for indian sign language recognition.
\newblock In \emph{ACM International Conference on Multimedia}, pages
  1366--1375, 2020.

\bibitem[Vasudevan et~al.(2020)Vasudevan, Negri, Linares-Barranco, and
  Serrano-Gotarredona]{vasudevan2020introduction}
Ajay Vasudevan, Pablo Negri, Bernabe Linares-Barranco, and Teresa
  Serrano-Gotarredona.
\newblock Introduction and analysis of an event-based sign language dataset.
\newblock In \emph{2020 15th IEEE International Conference on Automatic Face
  and Gesture Recognition (FG 2020)}, pages 675--682. IEEE, 2020.

\bibitem[von Agris and Kraiss(2010)]{von2010signum}
Ulrich von Agris and Karl-Friedrich Kraiss.
\newblock Signum database: Video corpus for signer-independent continuous sign
  language recognition.
\newblock In \emph{4th Workshop on the Representation and Processing of Sign
  Languages: Corpora and Sign Language Technologies}, pages 243--246, 2010.

\bibitem[Wan et~al.(2023)Wan, Mao, Zhang, and Dai]{wan2023rpeflow}
Zhexiong Wan, Yuxin Mao, Jing Zhang, and Yuchao Dai.
\newblock Rpeflow: Multimodal fusion of rgb-pointcloud-event for joint optical
  flow and scene flow estimation.
\newblock In \emph{Proceedings of the IEEE/CVF International Conference on
  Computer Vision}, pages 10030--10040, 2023.

\bibitem[Wang et~al.(2024{\natexlab{a}})Wang, Jin, Wu, Zhang, Zhu, Jiang, and
  Tian]{wang2024MvHeatDet}
Xiao Wang, Yu Jin, Wentao Wu, Wei Zhang, Lin Zhu, Bo Jiang, and Yonghong Tian.
\newblock Object detection using event camera: A moe heat conduction based
  detector and a new benchmark dataset.
\newblock \emph{arXiv preprint arXiv:2412.06647}, 2024{\natexlab{a}}.

\bibitem[Wang et~al.(2024{\natexlab{b}})Wang, Li, Zhu, Zhang, Chen, Li, Wang,
  Tian, and Wu]{wang2024visevent}
Xiao Wang, Jianing Li, Lin Zhu, Zhipeng Zhang, Zhe Chen, Xin Li, Yaowei Wang,
  Yonghong Tian, and Feng Wu.
\newblock Visevent: Reliable object tracking via collaboration of frame and
  event flows.
\newblock \emph{IEEE transactions on cybernetics}, 54\penalty0 (3):\penalty0
  1997--2010, 2024{\natexlab{b}}.

\bibitem[Wang et~al.(2024{\natexlab{c}})Wang, Li, Wang, Wang, Li, and
  Jiang]{wang2024r2gencsr}
Xiao Wang, Yuehang Li, Fuling Wang, Shiao Wang, Chuanfu Li, and Bo Jiang.
\newblock R2gencsr: Retrieving context samples for large language model based
  x-ray medical report generation.
\newblock \emph{arXiv preprint arXiv:2408.09743}, 2024{\natexlab{c}}.

\bibitem[Wang et~al.(2024{\natexlab{d}})Wang, Wang, Gao, Zhang, Wu, Xu, Shi,
  Wang, Li, Qian, et~al.]{wang2024searchingRAG}
Xiaohua Wang, Zhenghua Wang, Xuan Gao, Feiran Zhang, Yixin Wu, Zhibo Xu,
  Tianyuan Shi, Zhengyuan Wang, Shizheng Li, Qi Qian, et~al.
\newblock Searching for best practices in retrieval-augmented generation.
\newblock In \emph{Proceedings of the 2024 Conference on Empirical Methods in
  Natural Language Processing}, pages 17716--17736, 2024{\natexlab{d}}.

\bibitem[Wang et~al.(2025)Wang, Wang, Wang, Jiang, Li, Wang, Tian, and
  Tang]{wang2025ammrg}
Xiao Wang, Fuling Wang, Haowen Wang, Bo Jiang, Chuanfu Li, Yaowei Wang,
  Yonghong Tian, and Jin Tang.
\newblock Activating associative disease-aware vision token memory for
  llm-based x-ray report generation.
\newblock \emph{arXiv preprint arXiv:2501.03458}, 2025.

\bibitem[Wang et~al.(2021)Wang, Zhang, Wang, Wang, Huang, and
  Shen]{Wang2021Event-Based}
Yong Wang, Xian Zhang, Yanxiang Wang, Hongbin Wang, Chanying Huang, and Yiran
  Shen.
\newblock Event-based american sign language recognition using dynamic vision
  sensor.
\newblock In \emph{Workshop on Applications of Software Agents}, pages 3--10,
  2021.

\bibitem[Wu et~al.(2024)Wu, Gehrig, Lyu, Liu, and Gilitschenski]{wu2024leod}
Ziyi Wu, Mathias Gehrig, Qing Lyu, Xudong Liu, and Igor Gilitschenski.
\newblock Leod: Label-efficient object detection for event cameras.
\newblock In \emph{Proceedings of the IEEE/CVF Conference on Computer Vision
  and Pattern Recognition}, pages 16933--16943, 2024.

\bibitem[Yang et~al.(2023)Yang, Pan, and Liu]{yang2023event}
Yan Yang, Liyuan Pan, and Liu Liu.
\newblock Event camera data pre-training.
\newblock In \emph{Proceedings of the IEEE/CVF International Conference on
  Computer Vision}, pages 10699--10709, 2023.

\bibitem[Yao et~al.(2023)Yao, Zhou, Feng, Hu, Zhou, and Li]{yao2023IP-SLT}
Huijie Yao, Wengang Zhou, Hao Feng, Hezhen Hu, Hao Zhou, and Houqiang Li.
\newblock Sign language translation with iterative prototype.
\newblock In \emph{Proceedings of the IEEE/CVF International Conference on
  Computer Vision}, pages 15592--15601, 2023.

\bibitem[Ye et~al.(2023)Ye, Jiao, Wang, Tu, and Xiong]{ye2023cross}
Jinhui Ye, Wenxiang Jiao, Xing Wang, Zhaopeng Tu, and Hui Xiong.
\newblock Cross-modality data augmentation for end-to-end sign language
  translation.
\newblock \emph{arXiv preprint arXiv:2305.11096}, 2023.

\bibitem[Ye et~al.(2025)Ye, Wang, Jiao, Liang, and Xiong]{ye2025improving}
Jinhui Ye, Xing Wang, Wenxiang Jiao, Junwei Liang, and Hui Xiong.
\newblock Improving gloss-free sign language translation by reducing
  representation density.
\newblock \emph{Advances in Neural Information Processing Systems},
  37:\penalty0 107379--107402, 2025.

\bibitem[Yin et~al.(2022)Yin, Zhao, Jin, Zhang, Zeng, and He]{yin2022mlslt}
Aoxiong Yin, Zhou Zhao, Weike Jin, Meng Zhang, Xingshan Zeng, and Xiaofei He.
\newblock Mlslt: Towards multilingual sign language translation.
\newblock In \emph{Proceedings of the IEEE/CVF conference on computer vision
  and pattern recognition}, pages 5109--5119, 2022.

\bibitem[Yin et~al.(2023)Yin, Zhong, Tang, Jin, Jin, and Zhao]{yin2023gloss}
Aoxiong Yin, Tianyun Zhong, Li Tang, Weike Jin, Tao Jin, and Zhou Zhao.
\newblock Gloss attention for gloss-free sign language translation.
\newblock In \emph{Proceedings of the IEEE/CVF Conference on Computer Vision
  and Pattern Recognition}, pages 2551--2562, 2023.

\bibitem[Zhang et~al.(2025)Zhang, Tanzer, and Firat]{zhang2025scaling}
Biao Zhang, Garrett Tanzer, and Orhan Firat.
\newblock Scaling sign language translation.
\newblock \emph{Advances in Neural Information Processing Systems},
  37:\penalty0 114018--114047, 2025.

\bibitem[Zhang et~al.(2023)Zhang, Guo, Pan, Cai, Hong, Li, Yang, and
  Liu]{zhang2023remodiffuse}
Mingyuan Zhang, Xinying Guo, Liang Pan, Zhongang Cai, Fangzhou Hong, Huirong
  Li, Lei Yang, and Ziwei Liu.
\newblock Remodiffuse: Retrieval-augmented motion diffusion model.
\newblock In \emph{Proceedings of the IEEE/CVF International Conference on
  Computer Vision}, pages 364--373, 2023.

\bibitem[Zhang et~al.(2024)Zhang, Yin, Wang, Chen, Li, Wang, Lu,
  et~al.]{zhang2024evsign}
Pengyu Zhang, Hao Yin, Zeren Wang, Wenyue Chen, Shengming Li, Dong Wang,
  Huchuan Lu, et~al.
\newblock Evsign: Sign language recognition and translation with streaming
  events.
\newblock \emph{ECCV}, 2024.

\bibitem[Zhou et~al.(2023)Zhou, Chen, Clap{\'e}s, Wan, Liang, Escalera, Lei,
  and Zhang]{zhou2023GFSLT}
Benjia Zhou, Zhigang Chen, Albert Clap{\'e}s, Jun Wan, Yanyan Liang, Sergio
  Escalera, Zhen Lei, and Du Zhang.
\newblock Gloss-free sign language translation: Improving from visual-language
  pretraining.
\newblock In \emph{Proceedings of the IEEE/CVF International Conference on
  Computer Vision}, pages 20871--20881, 2023.

\bibitem[Zhou et~al.(2021)Zhou, Zhou, Qi, Pu, and Li]{zhou2021improving}
Hao Zhou, Wengang Zhou, Weizhen Qi, Junfu Pu, and Houqiang Li.
\newblock Improving sign language translation with monolingual data by sign
  back-translation.
\newblock In \emph{Proceedings of the IEEE/CVF Conference on Computer Vision
  and Pattern Recognition}, pages 1316--1325, 2021.

\bibitem[Zubic et~al.(2024)Zubic, Gehrig, and Scaramuzza]{zubic2024state}
Nikola Zubic, Mathias Gehrig, and Davide Scaramuzza.
\newblock State space models for event cameras.
\newblock In \emph{Proceedings of the IEEE/CVF Conference on Computer Vision
  and Pattern Recognition}, pages 5819--5828, 2024.

\end{thebibliography}
}

\end{document}